\pdfoutput=1
\documentclass[11pt]{article}

\usepackage[]{acl}

\usepackage{times}
\usepackage{latexsym}

\usepackage[T1]{fontenc}
\usepackage[utf8]{inputenc}

\usepackage{microtype}

\usepackage{booktabs}
\usepackage{times}
\usepackage{latexsym}
\usepackage{amsmath}
\usepackage{amssymb}
\usepackage{subcaption}
\usepackage{float}
\usepackage{multirow}
\usepackage{graphicx}
\usepackage{listings}
\usepackage{color, colortbl}

\usepackage{pifont}% http://ctan.org/pkg/pifont
\newcommand{\cmark}{\ding{51}}%
\newcommand{\xmark}{\ding{55}}%

\usepackage{xspace}
\usepackage[capitalise]{cleveref}
\usepackage{paralist}

\newcommand*{\eg}{e.g.\@\xspace}
\newcommand*{\ie}{i.e.\@\xspace}

\newcommand{\meddistant}{\textsc{MedDistant19}\xspace}

\makeatletter
\newcommand*{\etc}{%
    \@ifnextchar{.}%
        {etc}%
        {etc.\@\xspace}%
}
\makeatother

\usepackage{array}
\newcommand{\PreserveBackslash}[1]{\let\temp=\\#1\let\\=\temp}
\newcolumntype{C}[1]{>{\PreserveBackslash\centering}p{#1}}
\newcolumntype{R}[1]{>{\PreserveBackslash\raggedleft}p{#1}}
\newcolumntype{L}[1]{>{\PreserveBackslash\raggedright}p{#1}}

\usepackage{xcolor}

\colorlet{punct}{red!60!black}
\definecolor{background}{HTML}{EEEEEE}
\definecolor{delim}{RGB}{20,105,176}
\colorlet{numb}{magenta!60!black}

\lstdefinelanguage{json}{
    basicstyle=\normalfont\ttfamily,
    numbers=left,
    numberstyle=\scriptsize,
    stepnumber=1,
    numbersep=8pt,
    showstringspaces=false,
    breaklines=true,
    frame=lines,
    backgroundcolor=\color{background},
    literate=
     *{0}{{{\color{numb}0}}}{1}
      {1}{{{\color{numb}1}}}{1}
      {2}{{{\color{numb}2}}}{1}
      {3}{{{\color{numb}3}}}{1}
      {4}{{{\color{numb}4}}}{1}
      {5}{{{\color{numb}5}}}{1}
      {6}{{{\color{numb}6}}}{1}
      {7}{{{\color{numb}7}}}{1}
      {8}{{{\color{numb}8}}}{1}
      {9}{{{\color{numb}9}}}{1}
      {:}{{{\color{punct}{:}}}}{1}
      {,}{{{\color{punct}{,}}}}{1}
      {\{}{{{\color{delim}{\{}}}}{1}
      {\}}{{{\color{delim}{\}}}}}{1}
      {[}{{{\color{delim}{[}}}}{1}
      {]}{{{\color{delim}{]}}}}{1},
}
\definecolor{LightCyan}{rgb}{0.88,1,1}
\definecolor{Gray}{gray}{0.9}

\title{\meddistant: Towards an Accurate Benchmark for \\ Broad-Coverage Biomedical Relation Extraction}

\author{
  Saadullah Amin$^{\clubsuit,\triangle}$ \thanks{\quad \emph{Equal contribution.}} 
  \quad 
  Pasquale Minervini$^{\spadesuit\ *}$ 
  \quad 
  David Chang$^\diamondsuit$\\
  \quad 
  \textbf{Pontus Stenetorp}$^\spadesuit$ 
  \quad 
  \textbf{G\"unter Neumann}$^{\clubsuit,\triangle}$ \\
  $^\clubsuit$German Research Center for Artificial Intelligence 
  \quad 
  $^\spadesuit$UCL Centre for Artificial Intelligence \\
  $^\triangle$Saarland Informatics Campus, Saarland University 
  \quad
  $^\diamondsuit$Yale Center for Medical Informatics\\
   \small\texttt{\{saadullah.amin,guenter.neumann\}@dfki.de 
   \quad 
   \{p.minervini,p.stenetorp\}@cs.ucl.ac.uk}\\
  \small\texttt{david.chang@yale.edu}
}

\begin{document}

\maketitle

% =============================================================================
% ABSTRACT
% =============================================================================

\begin{abstract}
Relation extraction in the biomedical domain is challenging due to the lack of labeled data and high annotation costs, needing domain experts.
Distant supervision is commonly used to tackle the scarcity of annotated data by automatically pairing knowledge graph relationships with raw texts.
Such a pipeline is prone to noise and has added challenges to scale for covering a large number of biomedical concepts.
We investigated existing broad-coverage distantly supervised biomedical relation extraction benchmarks and found a significant overlap between training and test relationships ranging from 26\% to 86\%.
Furthermore, we noticed several inconsistencies in the data construction process of these benchmarks, and where there is no train-test leakage, the focus is on interactions between narrower entity types.
This work presents a more accurate benchmark \meddistant for broad-coverage distantly supervised biomedical relation extraction that addresses these shortcomings and is obtained by aligning the MEDLINE abstracts with the widely used SNOMED Clinical Terms knowledge base.
Lacking thorough evaluation with domain-specific language models, we also conduct experiments validating general domain relation extraction findings to biomedical relation extraction.
\end{abstract}

% -----------------------------------------------------------------------------

% =============================================================================
% SECTION 1: Introduction
% =============================================================================

\section{Introduction}
Extracting structured knowledge from unstructured text is important for knowledge discovery and management.
Biomedical literature and clinical narratives offer rich interactions between entities mentioned in the text~\cite{craven1999constructing,xu2014automatic}, which can be helpful for applications such as bio-molecular information extraction, pharmacogenomics, and identifying drug-drug interactions~(DDIs), among others~\cite{luo2017bridging}. 
Manually annotating these relations for training supervised learning systems is an expensive and time-consuming process \cite{segura20111st,kilicoglu2011constructing,segura2013semeval,li2016biocreative}, so the task often involves leveraging rule-based~\cite{abacha2011automatic,kilicoglu2020broad} and weakly supervised approaches~\cite{peng2016improving,dai2019distantly}.
%

% -----------------------------------------------------------------------------

% =============================================================================
% Figure 1.
% =============================================================================

\begin{figure}[!t]
    \includegraphics[width=1.0\linewidth]{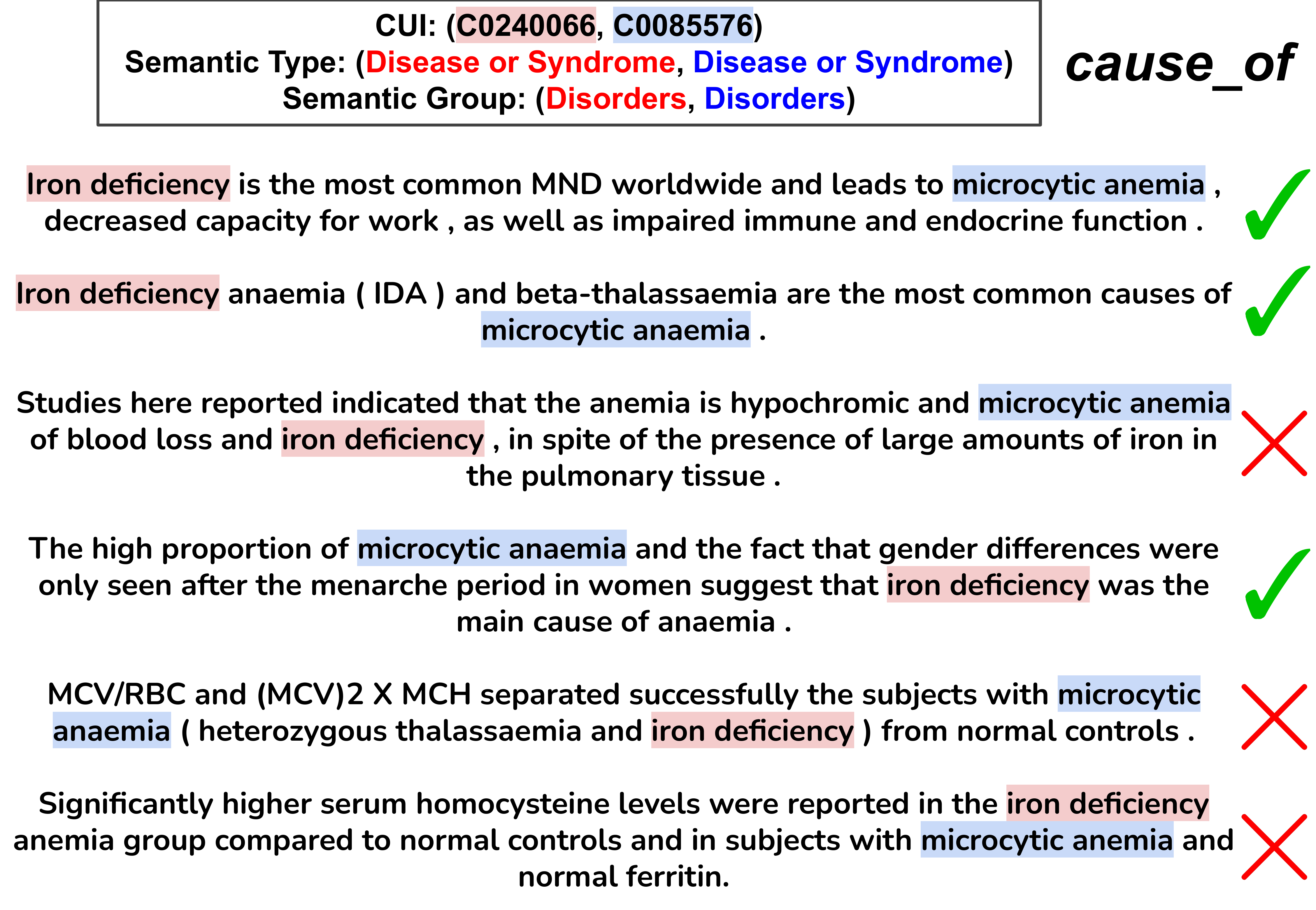}
    \centering
    \caption{
        An example of a bag instance representing the UMLS concept pair~(\texttt{C0240066}, \texttt{C0085576}) from the \meddistant dataset, expressing the relation \textit{cause\_of}.
        In this example, three out of six sentences express the relation, while others are incorrect labels resulting from the distant supervision.
    }
    \label{fig:bag_example}
\end{figure}

% -----------------------------------------------------------------------------

% =============================================================================
% Table 1
% =============================================================================

\begin{table*}[!t]
    \centering
    \resizebox{\linewidth}{!}{
    \begin{tabular}{rcccc}
        \toprule
        \textbf{Benchmark} & \textbf{Relations} & \textbf{No Train-Test Overlap} & \textbf{Broad-Coverage} & \textbf{Ontology} \\
        \midrule
        UMLS.v1 \cite{roller2014self} & 7 & - & \xmark & UMLS \\
        DTI \cite{hong-etal-2020-bere} & 6 & \cmark & \xmark & DrugBank \\
        UMLS.v2 \cite{amin2020data} & 355 & \xmark & \cmark & UMLS \\
        BioRel \cite{xing-etal-2020-biorel} & 125 & \xmark & \cmark & NDFRT, NCI \\
        UMLS.v3 \cite{hogan2021abstractified} & 275 & \xmark & \cmark & UMLS \\
        TBGA \cite{marchesin-silvello-2022} & 4 & \cmark & \xmark & DisGeNET \\
        \midrule
        \rowcolor{Gray}
        MedDistant19 & 22 & \cmark & \cmark & SNOMED CT \\
        \bottomrule
    \end{tabular}
    }
    \caption{
        The landscape of distantly supervised biomedical relation extraction~(Bio-DSRE) benchmarks: all the existing broad-coverage datasets have corpus-level triples overlap between the train and test splits~(\cref{table:leakage_stats}), where the knowledge graph~(KG) is also extracted from multiple ontologies. 
        The DTI and TBGA benchmarks focus on harmonized ontology but are limited to drug-target interactions and gene-disease associations. 
        In contrast, \meddistant has a broader coverage of entities and their semantic types and is normalized to a single ontology, SNOMED CT, which has significant clinical relevance. 
        We named the datasets from~\cite{roller2014self,amin2020data,hogan2021abstractified} to UMLS.v1/2/3 since the original works had no names. 
        For UMLS.v1, there is no publicly available code to reconstruct the dataset; thus, the overlap information is missing. 
    }\label{table:biods_bencmarks}
 \end{table*}
 
% -----------------------------------------------------------------------------

%
To scale to a large number of biomedical entities, recent works have focused on broad-coverage relation extraction \cite{amin2020data,xing-etal-2020-biorel,hogan2021abstractified}, where we investigated these benchmarks for possible train-test leakage of knowledge graph triples and found significant portions overlapping~(\cref{table:leakage_stats}).
Such leakage impacts the model performance as it allows to score higher by simply memorizing the training relations rather than generalizing to new, previously unknown ones.
We identify the sources of these issues as normalizing the textual form of concept mentions to their unique identifiers and improper handling of inverse relations.
In contrast, more accurate benchmarks exist \cite{hong-etal-2020-bere,marchesin-silvello-2022} but focus on narrower types of interactions.
To alleviate the broad-coverage benchmark issues and bridge this gap, we present a new benchmark \meddistant which draws its knowledge graph from the widely used healthcare ontology SNOMED CT \cite{chang-etal-2020-benchmark}. 
Further, with the success of domain-specific pre-trained language models for biomedical and clinical tasks \cite{gu2021domain}, and inspired by existing thorough relation extraction studies in the general domain \cite{peng2020learning,alt2020tacred,gao-etal-2021-manual}, we conduct an extensive evaluation using \meddistant for the biomedical domain.
% 

% =============================================================================
% Section 2: Related Work
% =============================================================================

\section{Related Work}
Relation Extraction~(RE) is an important task in biomedical applications. 
Traditionally, supervised methods require large-scale annotated corpora, which is impractical to scale for broad-coverage biomedical relation extraction~\cite{kilicoglu2011constructing,kilicoglu2020broad}. 
Distant Supervision~(DS) allows for the automated collection of noisy training examples by aligning a given knowledge base~(KB) with a collection of text sources \cite{mintz2009distant}.
DS was used in recent works \cite{alt2019fine,amin2020data} with pre-trained language models using Multi-Instance Learning~(MIL) by creating \emph{bags} of instances~\cite{riedel2010modeling} for corpus-level triple extraction.\footnote{RE is used to refer to two different tasks: sentence-level detection of relational instances and corpus-level triples extraction, a kind of knowledge graph completion or link prediction task \citep{pmlr-v119-amin20a}.}
In biomedical domain, \citet{roller2014self} first proposed the use of the Unified Medical Language System~(UMLS) Metathesaurus~\cite{bodenreider2004unified} as a KB with PubMed~\cite{canese2013pubmed} MEDLINE abstracts as text collection. 
For broad-coverage tasks, \citet{dai2019distantly} implemented a knowledge-based attention mechanism~\cite{han2018neural} for mutual learning with knowledge graph completion and entity type classification.
\citet{xing-etal-2020-biorel} introduced a large-scale BioRel benchmark focusing on drug-disease and gene-cancer interactions and showed significant performance using a comprehensive selection of baselines.
Recent works focused on using domain-specific pre-trained language models for distantly supervised biomedical relation extraction~(Bio-DSRE). 
\citet{amin2020data} extended relation enriched sentence-level BERT~\cite{wu2019enriching} to handle bag-level MIL and demonstrated that preserving the direction of the KB relationships can denoise the training signal.
They also outlined the steps to create a broad-coverage benchmark from UMLS.
Following this, \citet{hogan2021abstractified} introduced the concept of \emph{abstractified} MIL~(AMIL), by including different argument pairs belonging to the same semantic type pair in one bag, boosting performance on rare triples.
For domain-specific Bio-DSRE, \citet{hong-etal-2020-bere} introduced the BERE framework for latent tree learning and self-attention to use the semantic and syntactic information in the sentence for MIL. 
They also introduced a drug-target interactions (DTI) Bio-DSRE benchmark, suitable for drug repositioning, drawn from DrugBank \cite{wishart2018drugbank}.
Concurrent work of \citet{marchesin-silvello-2022} introduced a large-scale semi-automatically curated benchmark TGBA for gene-disease associations (GDA). TGBA uses DisGeNET \cite{pinero2020disgenet}, which collects data on human genotype-phenotype relationships. 
%

% =============================================================================
% TABLE 2
% =============================================================================

\begin{table}[!t]
   \centering
   \resizebox{\linewidth}{!}{
   \begin{tabular}{rccc}
       \toprule
       \textbf{\bf Triples} & \textbf{Train} & \textbf{Valid} & \textbf{Test} \\
       \midrule
       UMLS.v2 & 211,789 & 41,993 (26.7\%) & 89,486 (26.5\%) \\
       BioRel & 39,969 & 17,815 (86.17\%) & 17,927 (86.37\%) \\
       UMLS.v3 & 23,163 & 2,643 (44.38\%) & 5,184 (40.12\%) \\
       \bottomrule
   \end{tabular}
   }
   \caption{
        Training-test leakage we identified in the existing broad-coverage benchmarks.
        Numbers between parentheses show the percentage overlap of CUI triples.
   }
   \label{table:leakage_stats}
\end{table}
 
% -----------------------------------------------------------------------------

% =============================================================================
% Figure X.
% =============================================================================

\begin{figure*}
    \includegraphics[width=0.85\linewidth]{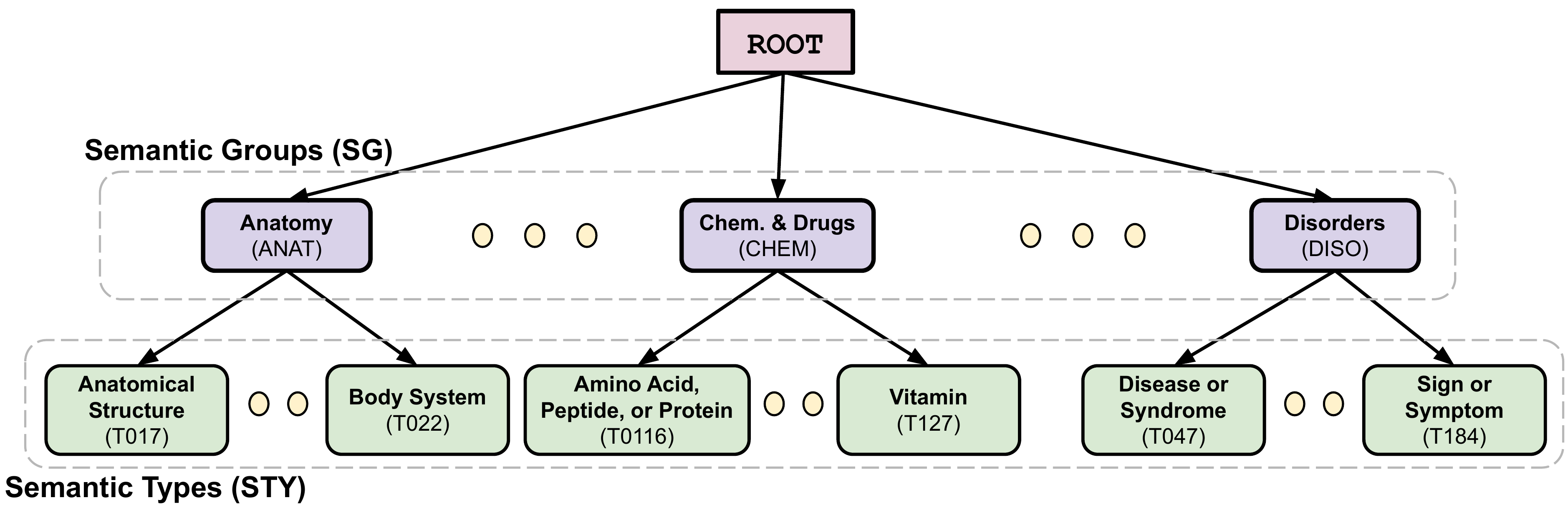}
    \centering
    \caption{
        Type hierarchy in UMLS, where each concept is classified under a taxonomy. 
        The \emph{coarse-grained} and \emph{fine-grained} entity types are referred to as Semantic Group~(SG) and Semantic Type~(STY) respectively.
    }
    \label{fig:ent_types}
\end{figure*}

% -----------------------------------------------------------------------------

%
This work investigates recent results from the broad-coverage Bio-DSRE literature by probing the respective datasets for overlaps between training and test sets.
Specifically, in UMLS, each concept is mapped to a \textit{Concept Unique Identifier (CUI)}, and a given CUI might have different surface forms~\cite{bodenreider2004unified}, we thus probe for CUI-based KG triples leakage.
Our results are shown in \cref{table:leakage_stats} for UMLS.v2 \cite{amin2020data}, BioRel \cite{xing-etal-2020-biorel}, and UMLS.v3 \cite{hogan2021abstractified}. 
%
%For UMLS.v2 and UMLS.v3, the main reason is to use textual variations of CUI as triples that result in the overlap.
%
For UMLS.v2 and UMLS.v3, the triples use surface forms of CUIs rather than the CUIs themselves, which results in an overlap between training and test sets.
For example, consider a relationship between a pair of UMLS entities (\texttt{C0013798}, \texttt{C0429028}).
These two entities can appear in different forms within a text, such as (\emph{electrocardiography}, \emph{Q-T interval}), (\emph{ECG}, \emph{Q-T interval}), and (\emph{EKG}, \emph{Q-T interval}); each of these distinct pairs still refers to the same original pair (\texttt{C0013798}, \texttt{C0429028}).
\citet{amin2020data} claim no such text-based leakage, but when canonicalized to their CUIs, this results in leakage across the splits as reported in \cref{table:leakage_stats}.
In contrast, BioRel directly splits CUI triples without accounting for inverse relations that can also result in leakage \cite{chang-etal-2020-benchmark}.
Since DSRE aims at corpus-level triples extraction, train-test triples leakage is problematic (see \cref{table:fixed_results}) compared to supervised sentence-level RE, where we aim to generalize %RE models 
to newer contexts.
We found no such overlap for DTI and TBGA, where the datasets used in \cite{roller2014self,dai2019distantly} are not publicly available.
Noting these shortcomings, we introduce a new and accurate benchmark \meddistant for broad-coverage Bio-DSRE. 
Our benchmark utilizes clinically relevant SNOMED CT Knowledge Graph~\cite{chang-etal-2020-benchmark}, extracted from the UMLS, that offers a careful selection of the concept types and is suitable for large-scale biomedical relation extraction.
\cref{table:biods_bencmarks} summarizes the current landscape of Bio-DSRE benchmarks.
In supervised RE, ChemProt \cite{krallinger2017overview} and DDI-2013 \cite{herrero2013ddi} focus on multi-class interactions between chemical-protein and drug-drug respectively. 
EU-ADR \cite{van2012eu} and GAD \cite{bravo2015extraction} focus on binary relations between genes and diseases, while CDR \cite{li2016biocreative} focuses on binary relations between chemicals and diseases.
%
%Therefore, being limited in their coverage of entity and relation types.
%

% -----------------------------------------------------------------------------

% =============================================================================
% TABLE 1
% =============================================================================

\begin{table}
    \centering
    \resizebox{\linewidth}{!}{
    \begin{tabular}{ccccc}
        \toprule
        \multirow{2}{*}{\bf Model and Data} & \multicolumn{2}{c}{\bf{Original}} & \multicolumn{2}{c}{\bf{Filtered}}\\
        & \textbf{AUC} & \textbf{F1} & \textbf{AUC} & \textbf{F1} \\
        
        \midrule
        \citet{amin2020data} & 68.4 & 64.9 & 50.8 & 53.1 \\
        %\citet{hogan2021abstractified} & 87.2 & 81.2 & - & - \\
        \citet{hogan2021abstractified}$^\dagger$ & 82.6 & 77.6 & 11.8 & 19.8 \\ 
        \bottomrule
    \end{tabular}
    }
    \caption{
        State-of-the-art Bio-DSRE language models were evaluated on the respective datasets before~(Original) and after~(Filtered) removing overlapping relationships. $^\dagger$ Our re-run of the AMIL (Type L) model; original scores are $87.2$ (AUC) and $81.2$ (F1).
    }
    \label{table:fixed_results}
 \end{table}

% -----------------------------------------------------------------------------

% =============================================================================
% SECTION 3: MedDistant19
% =============================================================================

\section{Constructing the MedDistant19 Benchmark} \label{sec:md19_construction}

\paragraph{Documents}
We used PubMed MEDLINE abstracts published up to 2019\footnote{\url{https://lhncbc.nlm.nih.gov/ii/information/MBR/Baselines/2019.html}} as our text source, containing 32,151,899 abstracts.
Following \citet{hogan2021abstractified}, we used \textsc{ScispaCy}~\footnote{\url{https://github.com/allenai/scispacy}}~\cite{neumann-etal-2019-scispacy} for sentence tokenization, resulting in 150,173,169 unique sentences.
We further introduce the use of \textsc{ScispaCy} for linking entity mentions to their UMLS CUIs and filtering disabled concepts from UMLS, which resulted in entity-linked mentions at the sentence-level.
Named entity recognition (NER) and normalization were two primary sources of errors in biomedical RE, as shown in \citet{kilicoglu2020broad}. 
While \textsc{ScispaCy} is reasonably performant among other options for biomedical entity linking, it remains quite noisy in practice; e.g., \citet{vashishth2021improving} showed that \textsc{ScispaCy} had only about a 50\% accuracy on extracting concepts in benchmark datasets. 
Despite this being a limitation, using %a concept extraction toolkit like 
\textsc{ScispaCy} is better than relying on string matching alone \cite{dai2019distantly,amin2020data,hogan2021abstractified}.
\paragraph{Knowledge Base}
We use UMLS2019AB~\footnote{\url{https://download.nlm.nih.gov/umls/kss/2019AB/umls-2019AB-full.zip}} as our primary knowledge source and apply a set of rules, resulting in a distilled and carefully reduced version of UMLS2019AB.
The UMLS Metathesaurus~\cite{bodenreider2004unified} covers concepts from 222 source vocabularies, thus being the most extensive ontology of biomedical concepts.
However, covering all ontologies can be challenging, given the interchangeable nature of the concepts.
For example, \emph{programmed cell death 1 ligand 1} is an alias of concept \texttt{C1540292} in the HUGO Gene Nomenclature Committee ontology~\citep{hgnc}, and it is an alias of concept \texttt{C3272500} in the National Cancer Institute Thesaurus.
This makes entity linking more challenging since a surface form can be linked to multiple entity identifiers and easier to have overlaps between training and test sets since the same fact may appear in both with different entity identifiers.
Furthermore, benchmark corpora for biomedical NER \cite{dougan2014ncbi,li2016biocreative} and RE~\cite{herrero2013ddi,krallinger2017overview} focuses on specific entity types~(\eg diseases, chemicals, proteins), and are usually normalized to a single ontology~\cite{kilicoglu2020broad}.
Following this trend, we also focus on a single vocabulary for Bio-DSRE.
We use SNOMED CT, the most widely used clinical terminology worldwide for documentation and reporting in healthcare~\cite{chang-etal-2020-benchmark}.
Since UMLS classifies each entity in a type taxonomy of semantic types (STY) and semantic groups (SG) (\cref{fig:ent_types}), this allows for narrowing the concepts of interest. 
Following \citet{chang-etal-2020-benchmark}, we first consider 8 semantic groups in SNOMED CT: Anatomy~(ANAT), Chemicals \& Drugs~(CHEM), Concepts \& Ideas~(CONC), Devices~(DEVI), Disorders~(DISO), Phenomena~(PHEN), Physiology~(PHYS), and Procedures~(PROC). 
We then remove CONC and PHEN as they are far too general to be informative for Bio-DSRE. 
For a complete list of semantic types covered in \meddistant, see \cref{table:md19_semantic_types_and_groups}.
Similarly, each relation is categorized into a type and has a reciprocal relation in UMLS (\cref{tab:rel_set_md19}), which can result in train-test leakage~\cite{dettmers2018convolutional}.
These steps follow \citet{chang-etal-2020-benchmark}, with the difference that we only consider relations of type \emph{has relationship other than synonymous, narrower, or broader}~(RO); this is consistent with prior works in Bio-DSRE.
We also exclude uninformative relations, \textit{same\_as}, \textit{possibly\_equivalent\_to}, \textit{associated\_with}, \textit{temporally\_related\_to}, and ignore inverse relations as generally is the case in RE. 
In addition, \citet{chang-etal-2020-benchmark} ensures that the validation and test set do not contain any new entities, making it a transductive learning setting where we assume all test entities are known beforehand.
However, we are expected to extract relations between unseen entities in real-world applications of biomedical RE.
To support this setup, we derive \meddistant using an inductive KG split method proposed by \citet{daza2021inductive} (see Appendix A in their paper).
\cref{table:snomed_kg_splits} summarizes the statistics of the KGs used for alignment with the text. 
We use split ratios of 70\%, 10\%, and 20\%.
Relationships are defined between CUIs and have no overlap between training, validation, and test. %set.
%

% -----------------------------------------------------------------------------

% =============================================================================
% Table 4
% =============================================================================

\begin{table}
    \centering
    \resizebox{\linewidth}{!}{
    \begin{tabular}{rcc}
        \toprule
        \textbf{Properties} & \textbf{Prior} & \textbf{MD19} \\
        \midrule
        \textit{approximate entity linking} & & \cmark \\
        \textit{unique NA sentences} & & \cmark \\
        \textit{inductive} & & \cmark \\
        \textit{triples leakage} & \cmark & \\
        \textit{NA-type constraint} & & \cmark \\
        \textit{NA-argument role constraint} & & \cmark \\
        \bottomrule
    \end{tabular}
    }
    \caption{
        \meddistant~(MD19) key data construction properties compared with the recent broad-coverage Bio-DSRE works.
        %
        %\emph{NA} represents relational sentence with \emph{unknown relation} type.
        %
    }\label{table:md19_props}
 \end{table}
 
% -----------------------------------------------------------------------------

% =============================================================================
% Table 2
% =============================================================================

\begin{table}[!t]
    \centering
    \resizebox{7.5cm}{!}{
    \begin{tabular}{rccc}
        \toprule
        {\bf Facts} & {\bf Training} & {\bf Validation} & {\bf Testing} \\
        \midrule
        Inductive &  261,797 & 48,641 & 97,861 \\
        Transductive & 318,524 & 28,370 & 56,812 \\
        \bottomrule
    \end{tabular}
    }
    \caption{
        The number of raw inductive and transductive SNOMED KG triples used for alignment with text.
        %
        %\meddistant uses the \emph{inductive} split. 
        %
        %See \cref{subsection:analysis} for comparison of the two.
    }\label{table:snomed_kg_splits}
 \end{table}
 
% -----------------------------------------------------------------------------

\subsection{Knowledge-to-Text Alignment} \label{section:knowledge_to_text}
We now describe the procedure for searching fact triples to match relational instances in text. 
Let $\mathcal{E}$ and $\mathcal{R}$ respectively denote the set of UMLS CUIs and relation types, and let $\mathcal{G} \subseteq \mathcal{E} \times \mathcal{R} \times \mathcal{E}$ denote the set of relationships contained in UMLS.
For producing a training-test split, we first create a set $\mathcal{G}^{+} \subseteq \mathcal{E} \times \mathcal{E}$ of related entity pairs as:
\begin{equation*}
    \mathcal{G}^{+} = \{ (e_{i}, e_{j}) \mid \langle e_{i}, p, e_{j} \rangle \in \mathcal{G} \lor \langle e_{j}, p, e_{i} \rangle \in \mathcal{G} \}
\end{equation*}
Following this, we obtain a set of unrelated entity pairs by corrupting one of the entities in each pair in $\mathcal{G}^{+}$ and making sure it does not appear in $\mathcal{G}^{+}$, obtaining a new set $\mathcal{G}^{-} \subseteq \mathcal{E} \times \mathcal{E}$ of unrelated entities, defined as follows:
\begin{equation*}
    \begin{aligned}
    \mathcal{G}^{-} & = \{ (\overline{e_{i}}, e_{j}) \mid (e_{i}, e_{j}) \in \mathcal{G}^{+} \land  (\overline{e_{i}}, e_{j}) \not\in \mathcal{G}^{+} \}\\
    & \cup \{ (e_{i}, \overline{e_{j}}) \mid (e_{i}, e_{j}) \in \mathcal{G}^{+} \land (e_{i}, \overline{e_{j}}) \not\in \mathcal{G}^{+} \}
    \end{aligned}
\end{equation*}
During the corruption process, we enforce two constraints:
\begin{inparaenum}[1)]
    \item \emph{type constraint} -- the two entities appearing in each negative pair in $\mathcal{G}^{-}$ should belong to an entity type pair from $\mathcal{G}^{+}$, and 
    \item \emph{role constraint} -- the noisy \emph{head} (\emph{tail}) entity in negative pair must have appeared in \emph{head} (\emph{tail}) role from a pair in $\mathcal{G}^{+}$. 
    %the entities used in the negative pair must have appeared in one or more positive pairs.
\end{inparaenum}
A naive choice for the negative group could be $\mathcal{G}^- = (\mathcal{E} \times \mathcal{E}) - \mathcal{G}^+$, for which the current approach is only a subset; however, enumerating all possible entity pairs can be infeasible if $|\mathcal{E}|$ is high.
Furthermore, we do not assume the completeness of UMLS, and only derive a \emph{fixed} sub-graph from the 2019 version subject to the constraints.
This process is similar to Local-Closed World Assumption~\cite[LCWA,][]{dong2014knowledge,DBLP:journals/pieee/Nickel0TG16}, in which a KG is assumed to be only locally complete: if we observed a triple for a specific entity $e_i \in \mathcal{E}$, then we assume that any non-existing relationship $(e_i, e_j)$ denotes a false fact and include them in $\mathcal{G}^{-}$.
Therefore, it is likely that if a triple emerges in a new PubMed article such that it violates the negative sampling assumptions, it will be considered a false negative.
However, this amount is negligible due to intractable search space that scales with the size of the KG. 
For each entity-linked sentence, we only consider those sentences that have SNOMED CT entities and have pairs in $\mathcal{G}^{+}$ and $\mathcal{G}^{-}$. 
Selected positive and negative pairs are mutually exclusive and have no overlap across splits. 
Since we only consider unique sentences associated with a pair, this makes for unique negative training instances, in contrast to \citet{amin2020data}, who considered generating positive and negative pairs from the same sentence. 
We define negative examples as relational sentences mentioning argument pairs with \emph{unknown relation type} (\texttt{NA}), \ie there might be a relationship, but the considered set of relations does not cover it. 
Our design choices are summarized in \cref{table:md19_props}. 
%

% =============================================================================
% Table 3
% =============================================================================

\begin{table}[!t]
    \centering
    \resizebox{\linewidth}{!}{
    \begin{tabular}{ccccccc}
        \toprule
        \multicolumn{2}{c}{\multirow{2}{*}{\textbf{Summary}}} & \textbf{Entities} & \textbf{Relations} & \textbf{STY} & \textbf{SG} \\
        \cline{3-6}
        & & 20,256 & 22 & 51 & 6 \\
        \midrule
        \textbf{Split} & \textbf{Instances} & \textbf{Facts} & \textbf{Bags} & \textbf{Inst. per Bag} & \textbf{NA (\%)} \\
        \midrule
        \textbf{Train} & 450,071 & 5,455 & 88,861 & 5.06 & 90.0\% \\
        \textbf{Valid} & 39,434 & 842 & 10,475 & 3.76 & 91.2\% \\
        \textbf{Test} & 91,568 & 1,663 & 22,606 & 4.05 & 91.1\% \\
        \bottomrule
    \end{tabular}
    }
    \caption{
        Summary statistics of the \meddistant dataset using Inductive SNOMED KG split (\cref{table:snomed_kg_splits}).
        The number of relations includes the unknown relation type~(\texttt{NA}).
    }
    \label{table:summary_stats}
 \end{table}

% -----------------------------------------------------------------------------

% =============================================================================
% Figure 2
% =============================================================================

\begin{figure}[!t]
    \includegraphics[width=1.0\linewidth]{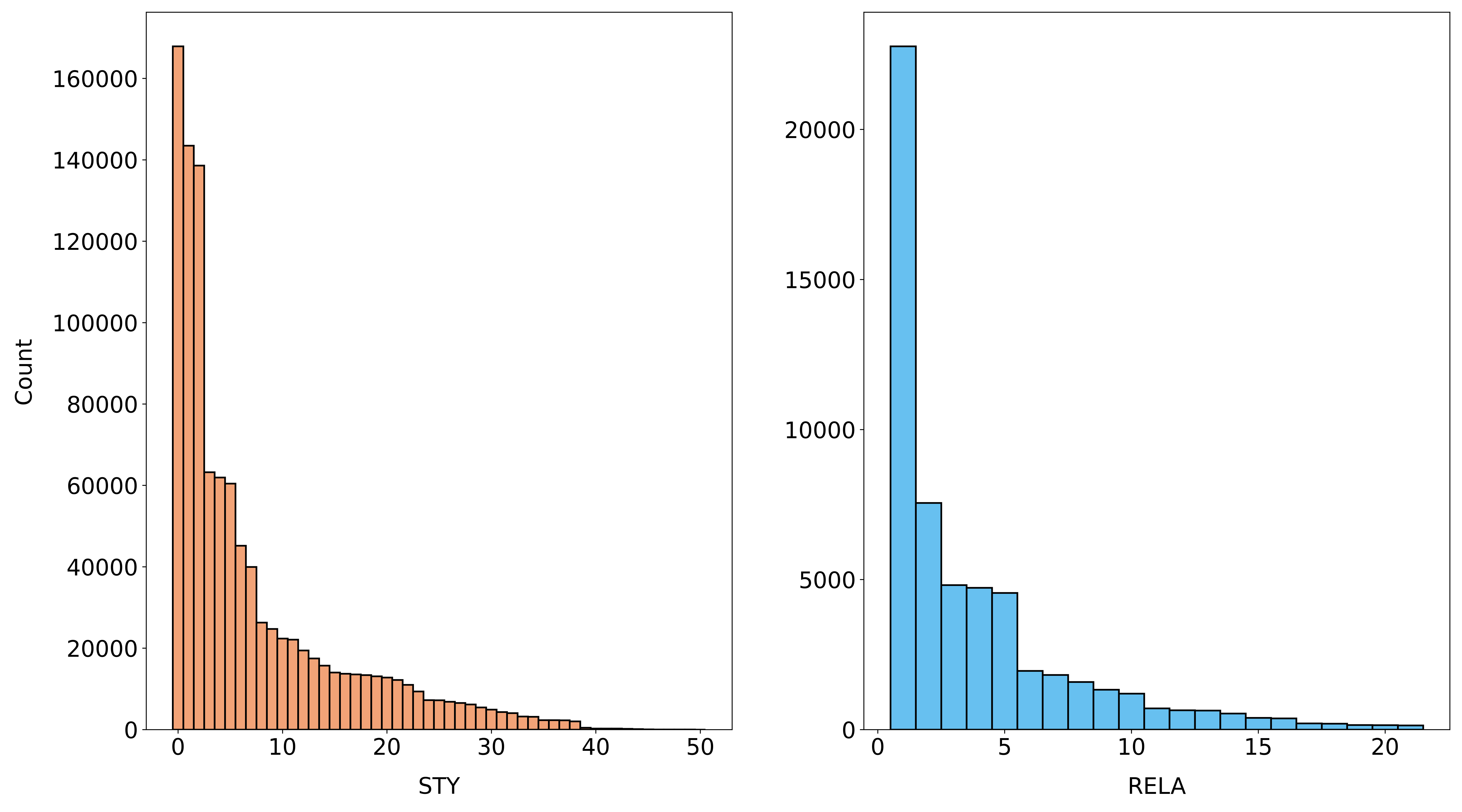}
    \centering
    \caption{
        (\textit{Left}) Entity distribution based on Semantic Types.
        (\textit{Right}) Relations distribution.
        }\label{fig:ent_rel_dist}
\end{figure}

% -----------------------------------------------------------------------------

% =============================================================================
% Table 4
% =============================================================================

\begin{table*}[!t]
    \centering
    \resizebox{0.9\linewidth}{!}{
    \begin{tabular}{ccccccccccc}
        \toprule
        \textbf{Model} & \textbf{Bag} & \textbf{Strategy} & \textbf{AUC} & \textbf{F1-micro} & \textbf{F1-macro} & \textbf{P@100} & \textbf{P@200} & \textbf{P@300} & \textbf{P@1k} & \textbf{P@2k} \\
        \midrule
        \multirow{5}{*}{CNN} & - & AVG & 27.3 & 33.0 & 16.1 & 50.0 & 46.0 & 44.0 & 41.0 & 33.6 \\
        & - & ONE & 30.4 & 36.7 & 18.2 & 67.0 & 58.5 & 52.6 & 43.5 & 34.4 \\
        \cmidrule{3-11}
        & \cmark & AVG & 30.4 & 36.2 & 19.8 & 70.0 & 58.0 & 56.0 & 46.0 & 35.5 \\
        & \cmark & ONE & 34.6 & 40.4 & 17.8 & 77.0 & 72.5 & 67.6 & 50.0 & 37.3 \\
        & \cmark & ATT & 35.0 & 40.1 & 19.8 & 78.0 & 73.5 & 68.6 & 51.4 & 36.4 \\
        \midrule
        \multirow{5}{*}{PCNN} & - & AVG & 27.2 & 32.4 & 12.9 & 54.0 & 49.5 & 50.3 & 40.7 & 33.2 \\
        & - & ONE & 29.8 & 36.7 & 16.2 & 66.0 & 55.5 & 52.3 & 44.4 & 34.2 \\
        \cmidrule{3-11}
        & \cmark & AVG & 29.6 & 37.3 & 20.5 & 59.0 & 50.5 & 50.0 & 47.0 & 35.9 \\
        & \cmark & ONE & 28.6 & 36.5 & 18.1 & 66.0 & 65.0 & 62.0 & 44.7 & 33.7 \\
        & \cmark & ATT & 32.5 & 38.2 & 14.4 & 71.0 & 71.0 & 67.3 & 49.0 & 35.2\\
        \midrule
        \multirow{5}{*}{GRU} & - & AVG & 42.7 & 47.4 & 27.8 & 78.0 & 74.0 & 76.0 & 59.2 & 42.7 \\
        & - & ONE & 46.4 & 49.3 & 29.2 & 86.0 & 80.5 & 78.3 & 61.2 & 44.9 \\
        \cmidrule{3-11}
        & \cmark & AVG & 28.6 & 37.2 & 17.9 & 57.0 & 57.0 & 56.0 & 45.3 & 35.4 \\
        & \cmark & ONE & 32.6 & 40.8 & 17.7 & 73.0 & 70.5 & 66.3 & 51.2 & 37.0 \\
        & \cmark & ATT & 36.6 & 40.9 & 22.2 & 77.0 & 72.0 & 67.6 & 51.3 & 38.7 \\
        \midrule
        \multirow{5}{*}{BERT} & - & AVG & \bf{79.8} & \bf{76.1} & \bf{65.3} & 95.0 & 96.0 & 96.0 & \bf{90.2} & 67.2 \\
        & - & ONE & 79.3 & \bf{76.1} & 64.7 & 93.0 & 94.0 & 94.0 & 89.2 & \bf{67.4} \\
        \cmidrule{3-11}
        & \cmark & AVG & 78.3 & 73.1 & 51.1 & \bf{99.0} & \bf{97.5} & \bf{96.6} & 87.8 & 66.0 \\
        & \cmark & ONE & 67.0 & 55.7 & 44.4 & 89.0 & 90.5 & 91.0 & 78.7 & 57.8 \\
        & \cmark & ATT & 64.6 & 56.4 & 42.7 & 89.0 & 87.5 & 85.6 & 75.4 & 57.9 \\
        \bottomrule
    \end{tabular}
    }
    \caption{
        Baseline results for \meddistant.
    }
    \label{table:main_results}
 \end{table*}
 
% -----------------------------------------------------------------------------

%
We also remove mention-level overlap across the splits and apply type-based mention pruning. 
Specifically, we pool mentions by type and remove the sentences which have the mention appearing more than 10,000 times.
We selected the threshold based on manual inspection of frequent mentions in each semantic type, so the information loss is minimal. 
At the same time, we still removed generalized mentions such as \emph{disease}, \emph{drugs}, \emph{temperature} etc.
We provide a complete list of mentions removed by this step in \cref{table:pruned_mentions}.
\cref{table:summary_stats} shows the final summary of \meddistant using inductive split covering 20,256 entities with 51 types and 343 type pairs. \cref{fig:ent_rel_dist} shows entity and relation plots, following a long-tail distribution.
%

% =============================================================================
% Section 4: Experiments
% =============================================================================

\section{Experiments}
\meddistant is released in a format that is compatible with the widely adopted RE framework OpenNRE \cite{han-etal-2019-opennre}.\footnote{\url{https://github.com/suamin/MedDistant19}}
To report our results, we use the \emph{corpus-level} Area Under the Precision-Recall (PR) curve (AUC), Micro-F1, Macro-F1, and Precision-at-$k$ (P@$k$) with $k \in \{ 100, 200, 300, 1k, 2k \}$, and the \emph{sentence-level} Precision, Recall, and F1.
Due to the imbalanced nature of relational instances, following \citet{gao-etal-2021-manual}, we report Macro-F1 values, and following \citet{hogan2021abstractified}, we report sentence-level RE results on relationships, including frequent and rare triples. 

\subsection{Baselines} \label{subsection:baselines}
Our baseline experiments largely follow the setup of \citet{gao-etal-2021-manual} with the addition of GRU models.\footnote{\url{https://github.com/pminervini/meddistant-baselines}}
For sentence encoding, we use CNN~\cite{liu2013convolution}, PCNN~\cite{zeng2015distant}, bidirectional GRU~\cite{hong-etal-2020-bere}, and BERT~\cite{devlin2019bert}.
We use GloVe~\cite{pennington2014glove} and Word2Vec~\cite{mikolov2013distributed} %\footnote{DRE baselines using CNN/PCNN models use 50-dimensional word embeddings from GloVe. Therefore, we trained 50-dim Word2Vec embeddings on PubMed abstracts.} 
for CNN/PCNN/GRU models and initialize BERT with BioBERT~\cite{lee2020biobert}.
We trained our models both at \emph{sentence-level} and at \emph{bag-level}.
In contrast, prior works only considered bag-level training for Bio-DSRE.
The sentence-level setup is similar to standard RE~\cite{wu2019enriching}, with the difference that the evaluation is conducted at the bag-level.
We also consider different pooling strategies, namely average (AVG), which averages the representations of sentences in a bag, at least one \citep[ONE,][]{zeng2015distant}, which generates relation scores for each sentence in a bag, and then selects the top-scoring sentence, and attention~(ATT), which learns an attention mechanism over the sentences within a bag.
%

% =============================================================================
% Figure 3
% =============================================================================

\begin{figure}[!t]
    \includegraphics[width=1\linewidth]{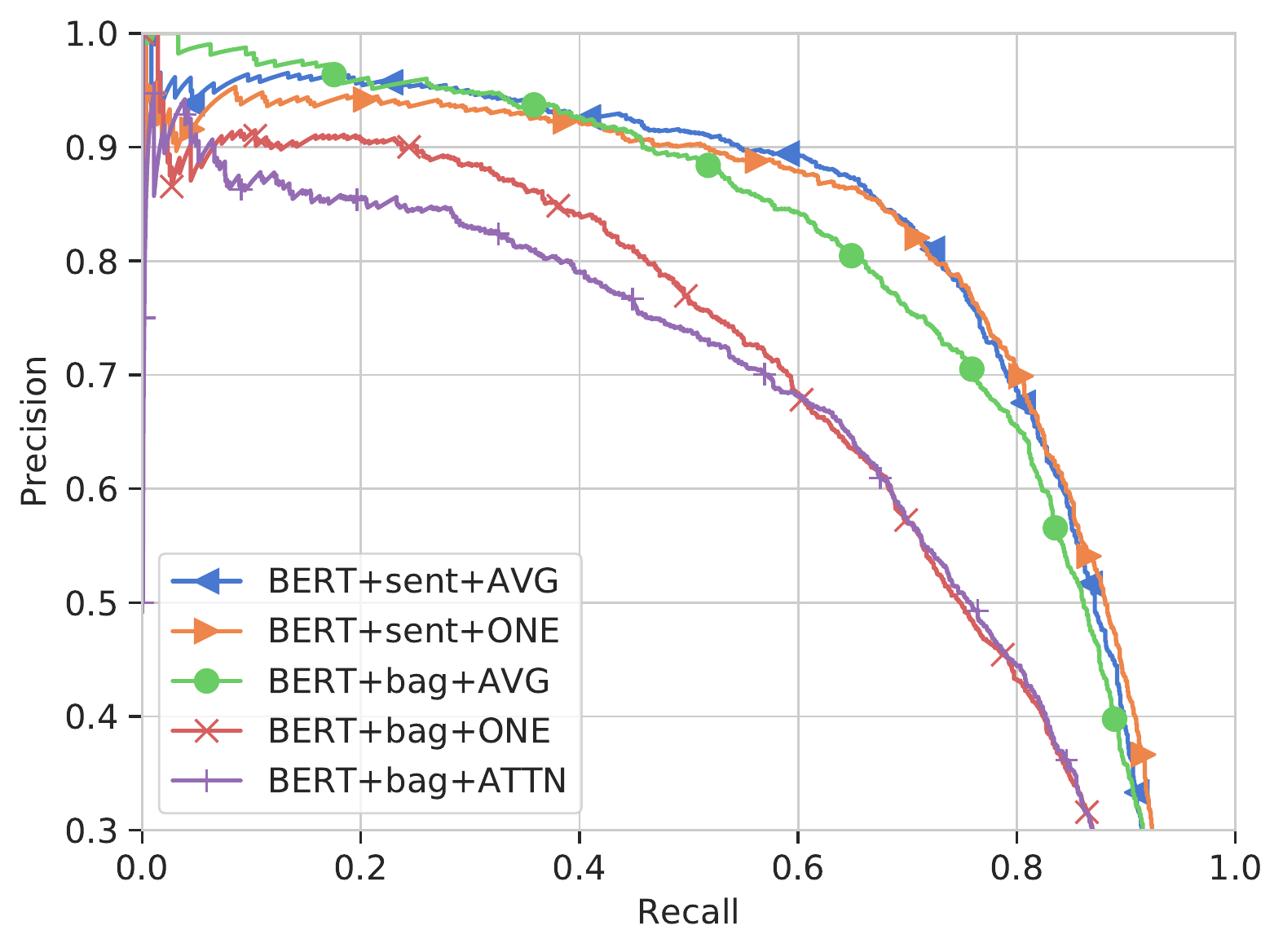}
    \centering
    \caption{
        Precision-Recall curves for BERT baselines.
        }
    \label{fig:pr_curves}
\end{figure}

% -----------------------------------------------------------------------------

% =============================================================================
% Table 3
% =============================================================================
 
\begin{table}[!t]
    \centering
    \resizebox{\linewidth}{!}{
    \begin{tabular}{rC{1.2cm}C{1.2cm}C{1.2cm}}
        \toprule
        \textbf{Model} & \textbf{1-1} & \textbf{1-M} & \textbf{M-1} \\
        \midrule
        BERT+bag+AVG & \bf{66.6} & \bf{48.3} & \bf{66.6} \\
        BERT+bag+ONE & 52.6 & 33.2 & 47.1 \\
        BERT+bag+ATT & 56.4 & 30.7 & 26.4 \\
        \bottomrule
    \end{tabular}
    }
    \caption{
        Averaged F1-micro score on relation-specific category for \emph{bag} pooling methods. The categories are defined using the \textit{cardinality} of head and tail SGs.
    }
    \label{table:sent_vs_bag_rel_cat}
 \end{table}
 
% -----------------------------------------------------------------------------

%
\cref{table:main_results} presents our main results. 
In all the cases, the BERT sentence encoder performed better than others since pre-trained language models are effective for entity-centric transfer learning~\cite{amin2021t2ner}, domain-specific fine-tuning \cite{amin2019mlt}, and can implicitly store relational knowledge during pre-training \cite{petroni2019language}. 
This trend is similar to the general domain, and the BERT-based experiments provide consistent baselines lacking in the prior works. 
Similar to the general domain \cite{gao-etal-2021-manual}, we find sentence-level training to perform better than the bag-level.
However, BERT+bag+AVG had much better precision for the top-scoring triples at the expense of long-tail performance.
At the sentence-level, those instances that have been correctly labeled by distant supervision~(\eg \cref{fig:bag_example}) provide enough learning signal, given the generalization abilities of LMs. 
However, the model is supposed to jointly learn from clean and noisy samples in bag-level training, thus limiting its overall performance. 
But, we do not find this trend for CNN/PCNN. 
Instead, the bag-level models performed slightly better except for GRU. 
We further plot Precision-Recall (PR) curves for BERT-based baselines in \cref{fig:pr_curves}.
\textbf{Pooling Strategies} In all cases, AVG proved to be a better pooling strategy; this finding is consistent with prior works.
Both \citet{amin2020data} and \citet{gao-etal-2021-manual} found ATT to produce less accurate results with LMs, which we also find to hold true for \meddistant.
To further study the impact of bag-level pooling strategies, we analyze the relation category-specific results.
Following \citet{chang-etal-2020-benchmark}, we grouped the relations based on cardinality, where the cardinality is defined as for a given relation type if the set of \emph{head} or \emph{tail} entities belongs to only one semantic group, then it has a cardinality one otherwise, M (many). 
The results are shown in \cref{table:sent_vs_bag_rel_cat} for bag-level BERT-based models with three pooling schemes. 
On average, models struggled the most with the 1-M category due to a lack of enough training signal to differentiate between heterogeneous entity types pooled over instances in a bag. 
While we would expect symmetric performance, to some extent, in 1-M and M-1 categories, the difference highlights that the KB-direction plays a role in Bio-DSRE, which previously has been used to de-noise the training signal \cite{amin2020data}. 
%

% =============================================================================
% Table 3
% =============================================================================

\begin{table}[!t]
    \centering
    \resizebox{\linewidth}{!}{
    \begin{tabular}{rC{1.2cm}C{1.2cm}C{1.2cm}}
        \toprule
        \textbf{Model} & \textbf{P} & \textbf{R} & \textbf{F1} \\
        \midrule
        \multicolumn{4}{c}{\bf All Triples}\\
        \midrule
        BERT+sent+AVG & \bf{0.79} & \bf{0.65} & \bf{0.71} \\
        BERT+bag+AVG & 0.72 & 0.64 & 0.68 \\
        \midrule
        \multicolumn{4}{c}{\bf Common Triples} \\
        \midrule
        BERT+sent+AVG & \bf{0.98} & \bf{0.62} & \bf{0.76} \\
        BERT+bag+AVG & 0.96 & 0.60 & 0.74 \\
        \midrule
        \multicolumn{4}{c}{\bf Rare Triples} \\
        \midrule
        BERT+sent+AVG & \bf{0.97} & 0.70 & 0.82 \\
        BERT+bag+AVG & 0.95 & \bf{0.73} & \bf{0.83} \\
        \bottomrule
    \end{tabular}
    }
    \caption{
        Sentence-level RE comparing BERT baselines trained at bag and sentence-level with AVG pooling on Rare and Common subsets of \meddistant.
        The triples include \texttt{NA} relational instances.
    }
    \label{table:sent_vs_bag_common_rare}
 \end{table}

% -----------------------------------------------------------------------------

%
\textbf{Long-Tail Performance} Following \citet{hogan2021abstractified}, we also perform sentence-level triples evaluation of BERT-based encoders trained at sentence-level and bag-level. 
The authors divided the triples (including \texttt{NA} instances) into two categories: those with 8 or more sentences are defined as \emph{common triples} and others as \emph{rare triples}.
\cref{table:sent_vs_bag_common_rare} shows these results. 
We note that both training strategies performed comparably on rare triples with BERT+sent+AVG more precise than BERT+bag+AVG at the expense of low recall. 
However, we find a noticeable difference in common triples where BERT+sent+AVG performed better. 
At the bag level, the model can overfit to certain type and mention heuristics, whereas sentence-level training allows more focus on context.
The current state-of-the-art model from \citet{hogan2021abstractified} creates a bag of instances by abstracting entity pairs belonging to the same semantic type pair into a single bag, thus producing heterogeneous bags. 
Due to such bag creation, it is not suited for sentence-level models.
%

% =============================================================================
% Figure 3
% =============================================================================

\begin{figure}[!t]
    \includegraphics[width=0.9\linewidth]{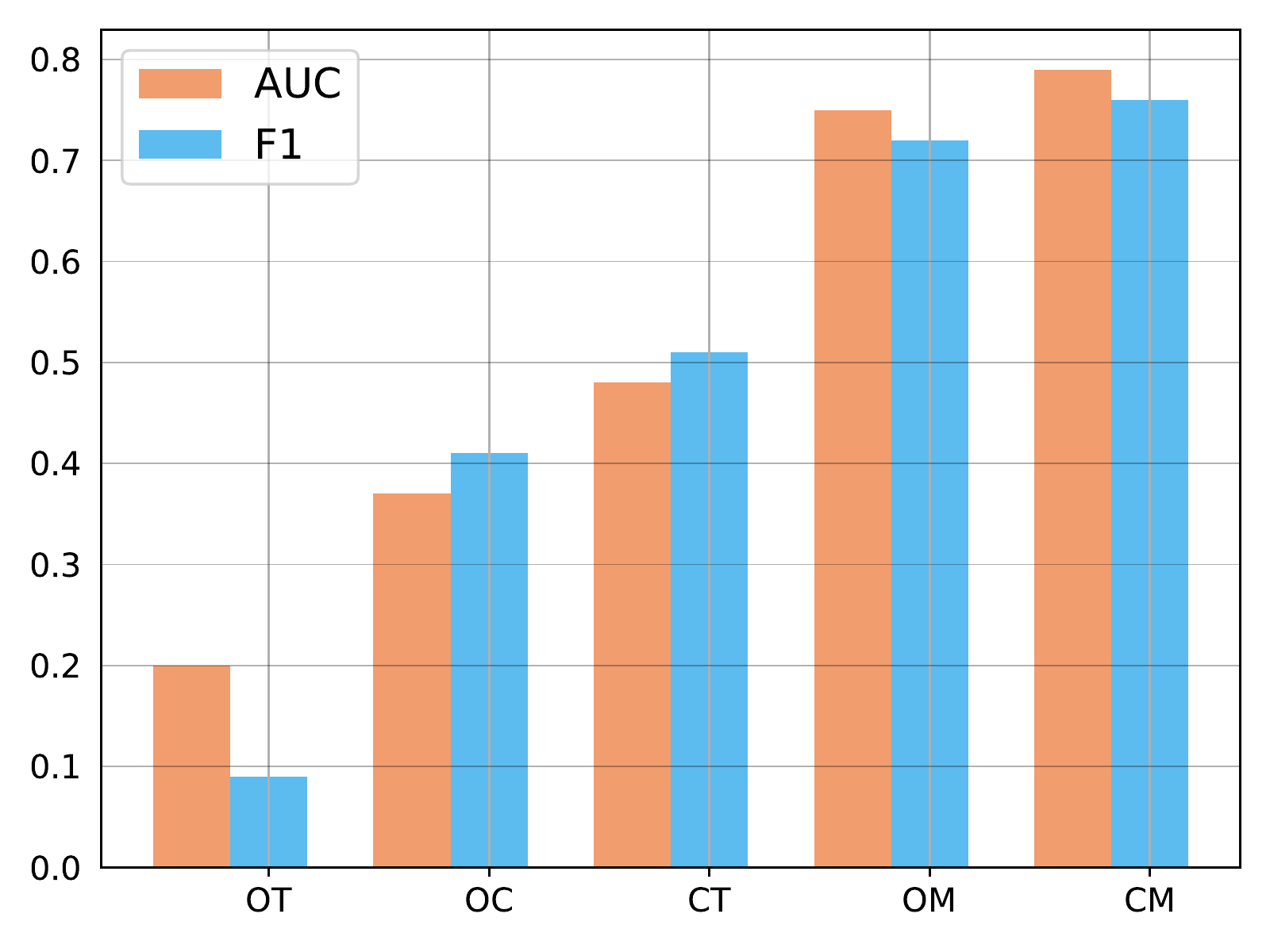}
    \centering
    \caption{Ablation showing the effect of different text encoding methods with \meddistant.}\label{fig:text_encoding_ablation}
\end{figure}

% -----------------------------------------------------------------------------

%
\subsection{Analysis} \label{subsection:analysis}

\textbf{Context, Mention, or Type?}
RE models are known to heavily rely on information from entity mentions, most of which is type information, and existing datasets may leak shallow heuristics via entity mentions that can inflate the prediction results \cite{peng2020learning}.
To study the importance of mentions, contexts, and entity types in \meddistant, we take inspiration from \cite{peng2020learning, han2020more} and conduct an ablation of different text encoding methods.
We consider entity mentions with special entity markers \cite{amin2020data} as the \emph{Context + Mention} (CM) setting, which is common in RE with LMs.
We then remove the context and only use mentions, the \emph{Only Mention} (OM) setting, which reduces to KG-BERT \cite{yao2019kg} for relation prediction.
We then only consider the context by replacing subject and object entities with special tokens, resulting in the \emph{Only Context} (OC) setting. 
Lastly, we consider two type-based (STY) variations as \emph{Only Type} (OT) and \emph{Context + Type} (CT). 
We train the models at the sentence-level and evaluate them at the bag-level.
We observe in \cref{fig:text_encoding_ablation} that the CM method had the highest performance, but surprisingly, OM performed quite well. 
This highlights the ability of LMs to memorize the facts and act as soft KBs \cite{petroni2019language}.
This trend is also consistent with general-domain \cite{peng2020learning}. 
The poor performance in the OC setting shows that the model struggles to understand the context, more pronounced in noise-prone distant RE than in supervised RE. 
Our CT setup can be seen as a sentence-level extrapolation of the AMIL model \cite{hogan2021abstractified}, which struggles to perform better than the baseline (OM).
However, comparing OC with CT, it is clear that the model benefits from type information as it can help constrain the space of the relations. 
Using only the type information had the least performance as the model fails to disambiguate between different entities belonging to the same type.
\textbf{Inductive or Transductive?} To study the impact of \emph{transductive} and \emph{inductive} splits (\cref{table:snomed_kg_splits}), we created another Bio-DSRE corpus using transductive train, validation, and test triples. 
The corpus generated differs from the inductive one, but it can offer insights into the model's ability to handle seen (\emph{transductive}) and unseen (\emph{inductive}) mentions. 
As shown in \cref{table:ind_vs_trans}, the performance using inductive is slightly better than transductive for corpus-level extractions in terms of AUC. However, the F1-macro score is better for transductive. 
We conclude that the model can learn patterns that exploit mentions and type information to extrapolate to unseen mentions in the inductive setup.
%

% =============================================================================
% Table 3
% =============================================================================
 
\begin{table}[!t]
    \centering
    \resizebox{\linewidth}{!}{
    \begin{tabular}{rccc}
        \toprule
        \textbf{Split} & \textbf{AUC} & \textbf{F1-micro} & \textbf{F1-macro} \\
        \midrule
        Inductive & \bf{79.9} & \bf{76.2} & 65.4 \\
        Transductive & 79.6 & 73.3 & \bf{65.9} \\
        \bottomrule
    \end{tabular}
    }
    \caption{
        BERT+sent+AVG performance on corpora created with an inductive and transductive set of triples.
    }
    \label{table:ind_vs_trans}
 \end{table}
 
% -----------------------------------------------------------------------------

\textbf{Does Expert Knowledge Help?} We now consider several pre-trained LMs with different knowledge capacities, specific to biomedical and clinical language understanding, to gain insights about the state-of-the-art encoders' performance and effectiveness on the \meddistant benchmark.
We use BERT \cite{devlin2019bert} as baseline.
We next consider only those pre-trained models trained with masked language modeling (MLM) objectives using domain-specific corpora. This includes ClinicalBERT \cite{alsentzer2019publicly}, BlueBERT \cite{peng2019transfer}, BioBERT \cite{lee2020biobert}, SciBERT \cite{beltagy2019scibert}, and PubMedBERT \cite{gu2021domain}.
We categorize these models as non-experts. %since they are only trained with Masked Language Modeling (MLM) objective.
Secondly, we consider expert models that modify the MLM objective or introduce new pre-training tasks using external knowledge, such as UMLS.
MedType \cite{vashishth2021improving}, initialized with BioBERT, is pre-trained to predict semantic types. 
KeBioLM \cite{yuan2021improving}, initialized with PubMedBERT, uses relational knowledge by initializing the entity embeddings with TransE \cite{bordes2013translating}, improving entity-centric tasks, including RE. 
UmlsBERT \cite{michalopoulos2021umlsbert}, initialized with ClinicalBERT, modifies MLM to mask words belonging to the same CUI and further introduces semantic type embeddings. SapBERT \cite{liu2021self}, initialized with PubMedBERT, introduces a metric learning task for clustering synonyms together in an embedding space.
%

% =============================================================================
% Table 11
% =============================================================================

\begin{table}
    \centering
    \resizebox{\linewidth}{!}{
    \begin{tabular}{rcccccc}
        \toprule
        \multirow{2}{*}{\textbf{Encoder}} & \multicolumn{5}{c}{\textbf{Knowledge Type}} & \multirow{2}{*}{\textbf{AUC}} \\
        \cline{2-6}
        & \textit{Biomedical} & \textit{Clinical} & \textit{Type} & \textit{Triples} & \textit{Synonyms} \\
        \midrule
        & \multicolumn{5}{c}{\textsc{Non-Expert Models}} & \\
        \midrule
        BERT & & & & & & 0.72 \\
        ClinicalBERT & \cmark & \cmark & & & & 0.73 \\
        BlueBERT & \cmark & & & & & 0.78 \\
        SciBERT & \cmark & & & & & 0.78 \\
        BioBERT & \cmark & & & & & 0.79 \\
        PubMedBERT & \cmark & & & & & \textbf{0.80} \\
        \midrule
        & \multicolumn{5}{c}{\textsc{Expert Knowledge Models}} & \\
        \midrule
        MedType & \cmark & & \cmark & & & 0.77 \\
        KeBioLM & \cmark & & & \cmark & & \textbf{0.80} \\
        UmlsBERT & \cmark & \cmark & \cmark & &  & 0.75 \\
        SapBERT & \cmark & & & & \cmark & 0.78 \\
        \bottomrule
    \end{tabular}
    }
    \caption{
        Expert and non-expert pre-trained language models performance on \meddistant.
    }\label{table:encoders_study}
 \end{table}
 
% -----------------------------------------------------------------------------

%
\cref{table:encoders_study} shows the results of these sentence encoders fine-tuned on the \meddistant dataset at sentence-level with AVG pooling. 
Without domain-specific knowledge, BERT performs slightly worse than the lowest-performing biomedical model, highlighting the presence of shallow heuristics in the data common to the general and biomedical domains. 
While domain-specific pre-training improves the results, similar to \citet{gu2021domain}, we find clinical LMs underperform on the biomedical RE task. 
There was no performance gap between BlueBERT, SciBERT, and BioBERT. However, PubMedBERT brought improvement, consistent with \citet{gu2021domain}. 
For expert knowledge-based models, we noted a negative impact on performance.
While we would expect type-based models, MedType and UmlsBERT, to bring improvement, their effect can be attributed to overfitting certain types and patterns. 
KeBioLM, initialized with PubMedBERT, has the same performance despite seeing the triples used in \meddistant during pre-training, highlighting the difficulty of the Bio-DSRE. SapBERT, which uses the knowledge of synonyms, also hurt PubMedBERT's performance, suggesting that while synonyms can help in entity linking, RE is a more challenging task in noisy real-world scenarios.
%

% =============================================================================
% SECTION 5
% =============================================================================

\section{Discussion}
In the biomedical domain, health experts are often concerned with a particular type of interaction, for example, drug-target and gene-disease. 
However, the number of ontologies is constantly growing (222 in UMLS2019AB), thus a growing need for a more general purpose relation extraction benchmark. 
Broad-coverage benchmarks exist for biomedical entity linking, such as MedMentions \cite{mohan2018medmentions}, but they still lack many important concepts involved in relational learning. 
The research community has come up with several RE benchmarks (see \cref{table:biods_bencmarks}), but the challenge remains as new entities, and relations emerge with the constant growth of biomedical literature.
Hence, constructing a broad benchmark for biomedical RE is challenging due to domain requirements; nonetheless, having an accurate benchmark could offer a utility for future research. 
We supplement this discussion with \cref{appendix_section:limitations} for a note on limitations.
Further, the train-test overlap highlights the need to systematically assess the proposed benchmarks for inconsistencies that can overestimate the model performance.
Similar assessments have shown up in QA generalization where train-test overlap inflates the model performance \cite{liu-etal-2022-challenges}.
Related to RE generalization, \citet{rosenman-etal-2020-exposing} exposed shallow heuristics while \citet{taille-etal-2021-separating} showed that neural RE models could retain triples, primarily due to type hints.
\meddistant partially addresses these issues by an inductive setup that can offer insights into the generalization trend in biomedical RE using unseen entities. 
%

% =============================================================================
% SECTION 6
% =============================================================================

\section{Conclusion}
In this work, we highlighted a need for an accurate broad-coverage benchmark for Bio-DSRE.
We bridged this gap by utilizing SNOMED CT for constructing the benchmark and laying out the best practices.
We thoroughly evaluated the benchmark with baselines and state-of-the-art, showing there is room to conduct further research.
%

% =============================================================================
% ACKNOWLEDGMENTS
% =============================================================================

\section*{Acknowledgments}
The authors would like to thank the anonymous reviewers for their helpful feedback and William Hogan for assistance in providing data and code for AMIL experiments.
The work was partially funded by the European Union (EU) Horizon 2020 research and innovation program through the projects Precise4Q (777107) and Clarify (875160), and the German Federal Ministry of Education and Research (BMBF) through the projects CoRA4NLP (01IW20010) and XAINES (01IW20005).
The authors also acknowledge the computing resources provided by the DFKI and UCL.

\section*{Legal \& Ethical Considerations}

\textbf{Does the dataset contain information that might be considered sensitive or confidential? (\eg personally identifying information)} We use PubMed MEDLINE abstracts \cite{canese2013pubmed}\footnote{\url{https://lhncbc.nlm.nih.gov/ii/information/MBR/Baselines/2019.html}} that are publicly available and is distributed by National Library of Medicine (NLM). 
These texts are in the biomedical and clinical domains and are almost entirely in English. It is standard to use this corpus as a text source in several biomedical LMs \cite{gu2021domain}. 
We cannot claim the guarantee that it does not contain any confidential or sensitive information e.g, it has clinical findings mentioned throughout the abstracts such as \textit{A twenty-six-year-old male presented with high-grade fever}, which identifies the age and gender of a patient but not the identity. 
We did not perform a thorough analysis to distill such information since it is in the public domain. 
%

% =============================================================================
% BIBLIOGRAPHY
% =============================================================================

\bibliography{custom}
\bibliographystyle{acl_natbib}

% =============================================================================
% APPENDIX
% =============================================================================

\clearpage 
\appendix

% Appendix tables, figures, equations numbering
\setcounter{table}{0}
\renewcommand{\thetable}{A.\arabic{table}}

\setcounter{figure}{0}
\renewcommand{\thefigure}{A.\arabic{figure}}

\section{UMLS}

This section presents additional details about UMLS, including the final set of relations considered in \meddistant (with their inverses obtained from the UMLS) and a complete list of semantic types (STY). 
Since, in relation extraction (RE), we are not interested in bidirectional extractions, therefore it is sufficient to only model one direction. 
Previous studies \cite{xing-etal-2020-biorel,amin2020data,hogan2021abstractified} fail to account the inverse relations, and with naive split, it can lead to train-test leakages. 
For more discussion on the relations in UMLS, including transitive closures, see Section 3.1 in \citet{chang-etal-2020-benchmark}.
We used UMLS2019AB to be consistent with the prior works.

\subsection{UMLS Files}

In UMLS \cite{bodenreider2004unified}, a concept is provided with a unique identifier called Concept Unique Identifier (CUI), a term status (TS), and whether or not the term is preferred (TTY) in a given vocabulary, e.g., SNOMED CT. 
The concepts are stored in a file distributed by UMLS called \texttt{MRCONSO.RRF}.\footnote{\url{https://www.ncbi.nlm.nih.gov/books/NBK9685/table/ch03.T.concept_names_and_sources_file_mr/}} Each concept further belongs to one or more semantic types (STY), provided in a file called \texttt{MRSTY.RRF}, with a type identifier TUI. There are 127 STY\footnote{\url{https://lhncbc.nlm.nih.gov/ii/tools/MetaMap/Docs/SemanticTypes_2018AB.txt}} in the UMLS2019AB version, which are mapped to 15 semantic groups (SG).\footnote{\url{https://lhncbc.nlm.nih.gov/ii/tools/MetaMap/Docs/SemGroups_2018.txt}}. 
The relationships between the concepts are organized in a multi-relational graph distributed in a file called \texttt{MRREL.RRF}\footnote{\url{https://www.ncbi.nlm.nih.gov/books/NBK9685/table/ch03.T.related_concepts_file_mrrel_rrf/?report=objectonly}}. 
The final set of relations considered in \meddistant is presented in \cref{tab:rel_set_md19}. 
Note that we only consider relations belonging to the \emph{RO} (\emph{has a relationship other than synonymous, narrower, or broader}) type, which is consistent with prior works. 
This consideration ignores relations such as \emph{isa}, which defines hierarchy among relations.
%

% =============================================================================
% Figure 1
% =============================================================================

\begin{figure}
    \includegraphics[width=0.8\linewidth]{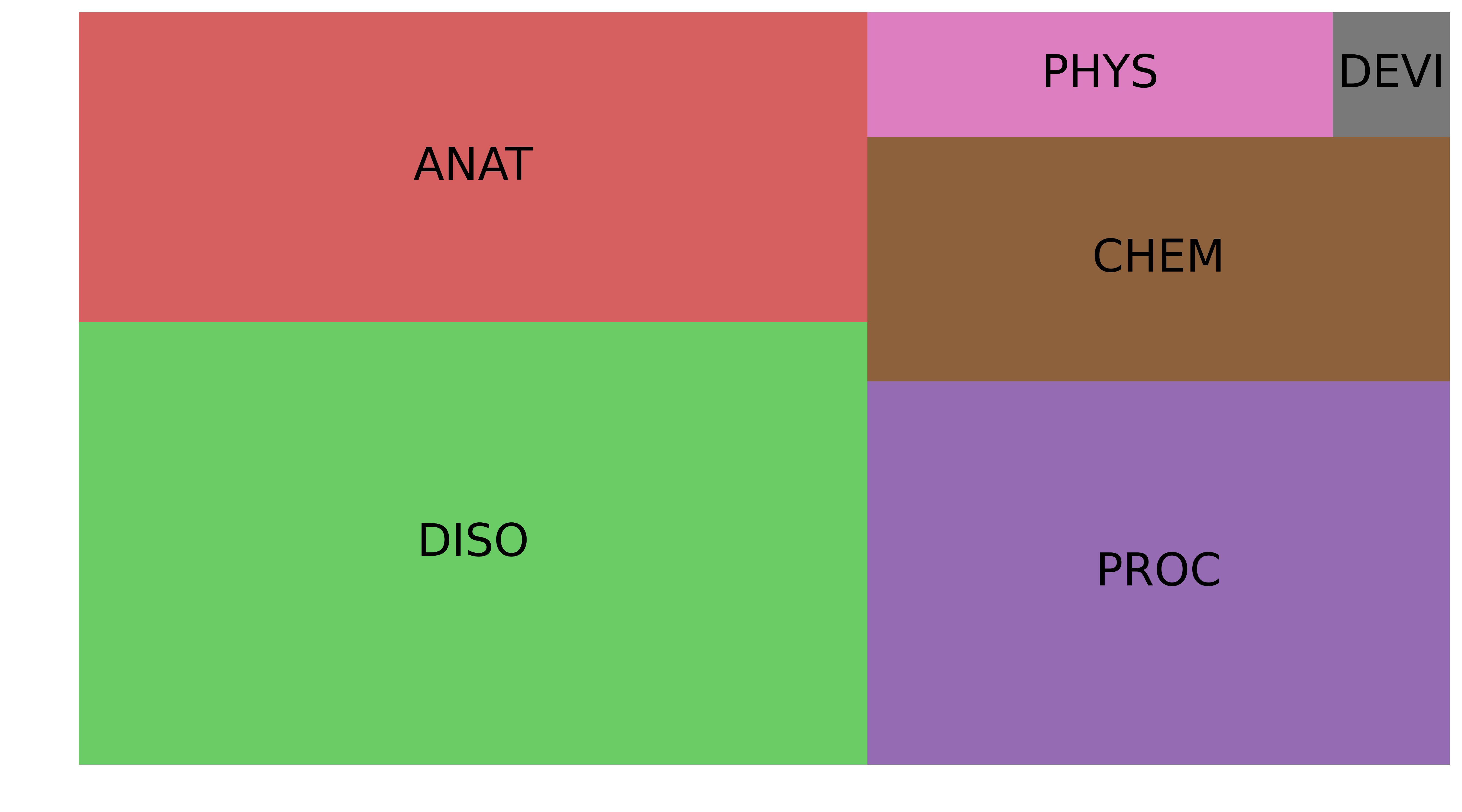}
    \centering
    \caption{
        Relative proportions of the entities present in \meddistant, based on the semantic groups.
    }
    \label{fig:sg_relative_prop}
\end{figure}

% -----------------------------------------------------------------------------

\subsection{Semantic Groups and Semantic Types}

As we noted in \cref{fig:ent_rel_dist}, entities and relations follow a long-tail distribution. 
This has a major impact on the quality of the dataset created. 
For example, in the general domain, the standard benchmark NYT10~\cite{riedel2010modeling} has more than half of the positive instances belonging to one relation type \texttt{/location/location/contains}.
\cref{fig:sg_relative_prop} shows the relative proportions of the semantic groups in \meddistant.
Further, we used an inductive split set with 70, 10, and 20 proportions of train, validation, and test splits for constructing \meddistant.
Below is an example instance from the dataset in OpenNRE \cite{han-etal-2019-opennre} format:\\

\begin{lstlisting}[language=json,numbers=none,backgroundcolor=\color{white}]
{
  "text": "In one patient who 
  showed an increase of plasma 
  prolactin level , associated 
  with low testosterone and 
  LH , a microadenoma 
  of the pituitary gland 
  ( prolactinoma ) was 
  detected .", 
  "h": {
    "id": "C0032005", 
    "pos": [130, 145], 
    "name": "pituitary gland"
  }, 
  "t": {
    "id": "C0033375", 
    "pos": [148, 160], 
    "name": "prolactinoma"
  }, 
  "relation": "finding_site_of"
}

/-----------------------------/

{
  "text": "Severe heart disease
  may result in cardiac cirrhosis
  in the elderly , with ascites
  and hepatomegaly .", 
  "h": {
    "id": "C0018799",
    "pos": [7, 20],
    "name": "heart disease"
  },
  "t": {
    "id": "C0085699",
    "pos": [35, 52],
    "name": "cardiac cirrhosis"
  },
  "relation": "cause_of"
}

/-----------------------------/

{
  "text": "Complications 
  closely associated to the 
  osteosynthesis appeared 
  only in instable 
  fractures ( 7 % ) .",
  "h": {
    "id": "C0016658",
    "pos": [81, 90],
    "name": "fractures"
  },
  "t": {
    "id": "C0016642",
    "pos": [40, 54],
    "name": "osteosynthesis"
   },
   "relation": 
   "direct_morphology_of"
}

/-----------------------------/

{
  "text": "Gluten proteins ,
  the culprits in celiac 
  disease ( CD ) , show 
  striking similarities in 
  primary structure with 
  human salivary proline-rich 
  proteins ( PRPs ) .",
  "h": {
    "id": "C2362561",
    "pos": [0, 15],
    "name": "Gluten proteins"
  },
  "t": {
    "id": "C0007570",
    "pos": [34, 48],
    "name": "celiac disease"
  },
  "relation": 
  "causative_agent_of"
}

/-----------------------------/

{
  "text": "Postherpetic
  neuralgia is an unfortunate
  aftermath of shingles ,
  and is most likely to
  develop , and most
  persistent , in elderly
  patients .", 
  "h": {
    "id": "C0032768",
    "pos": [0, 22],
    "name": "Postherpetic 
    neuralgia"
  }, 
  "t": {
    "id": "C0019360",
    "pos": [54, 62],
    "name": "shingles"
  },
  "relation": "occurs_after"
}
\end{lstlisting}

\section{UMLS License Agreement}

To use the \meddistant benchmark, the user must have signed the UMLS agreement\footnote{\url{https://uts.nlm.nih.gov/license.html}}. 
The UMLS agreement requires those who use the UMLS \cite{bodenreider2004unified} to file a brief report once a year to summarize their use of the UMLS. 
It also requires acknowledging that the UMLS contains copyrighted material and that those restrictions are respected. 
The UMLS agreement requires users to agree to obtain agreements for \emph{each} copyrighted source before its use within a commercial or production application.

\section{Risks}

While our work does not have direct risk, we provide the dataset while asking users to respect the UMLS license before downloading it. 
This user agreement is needed to use our benchmark and to respect the source ontologies licenses. 
We provide this with the hope to accelerate reproducible research in Bio-DSRE by having ready-to-use corpora, with only the condition that the user has obtained the license. 
We provide users with this note and hope this will be respected. 
However, there is a risk that users may download the data and re-distribute it without respecting the UMLS license. 
In case of such exploitation, we will add the UMLS authentication layer to protect data, where the user will be required to provide a UMLS API key, which will be validated, and only then will the data be allowed to be downloaded.
%

% =============================================================================
% SECTION X
% =============================================================================

\section{Limitations} \label{appendix_section:limitations}
We provide several limitations of our work as presented in its current form. 
\meddistant aims to introduce a new benchmark with good practices. However, it is still limited in its scope of ontologies considered. 
It also has a limited subset of relation types provided by UMLS. 
For example, the current benchmark does not include an important relation \textit{may\_treat}, because it is outside SNOMED CT. 
Since \meddistant is focused on SNOMED CT, it lacks coverage of important protein-protein interactions, drug side-effects, and relations involving genes as provided by RxNorm \cite{nelson2011rxnorm}, Gene Ontology \cite{10.1093/nar/gky1055}, etc. 
\meddistant is automatically-created and susceptible to noise and thus needs to be approached carefully as a potential source for biomedical knowledge. 
While the dataset was not created to represent \emph{true} biomedical knowledge, it has the potential to be treated as a reliable reference. 

\begin{table}[!t]
    \centering
    \resizebox{7.5cm}{!}{
    \begin{tabular}{rccc}
        \toprule
        {\bf Encoder} & {\bf Bag Size} & {\bf Batch Size} & {\bf Embedding} \\
        \midrule
        CNN+sent+AVG & - & 128 & biowordvec \\
        CNN+sent+ONE & - & 128 & biowordvec \\
        CNN+bag+AVG & 8 & 128 & GloVe \\
        CNN+bag+ONE & 16 & 256 & GloVe \\
        CNN+bag+ATT & 8 & 256 & GloVe \\
        \midrule
        PCNN+sent+AVG & - & 128 & biowordvec \\
        PCNN+sent+ONE & - & 128 & biowordvec \\
        PCNN+bag+AVG & 4 & 128 & GloVe \\
        PCNN+bag+ONE & 8 & 128 & GloVe \\
        PCNN+bag+ATT & 8 & 128 & GloVe \\
        \midrule
        GRU+sent+AVG & - & 128 & biowordvec \\
        GRU+sent+ONE & - & 128 & biowordvec \\
        GRU+bag+AVG & 8 & 128 & biow2v \\
        GRU+bag+ONE & 16 & 256 & GloVe \\
        GRU+bag+ATT & 16 & 128 & GloVe \\
        \bottomrule
    \end{tabular}
    }
    \caption{
        Best hyperparameters for CNN, PCNN, and GRU sentence encoders.
    }\label{table:hyperparams_cnn}
 \end{table}
 
\section{Experimental Setup and Hyperparameters}

We followed the experimental setup of \citet{gao-etal-2021-manual} for BERT-based experiments.
Specifically, we used batch size 64, with a learning rate of 2e-5, maximum sequence length 128, and bag size 4. We used a single NVIDIA Tesla V100-32GB for BERT-based experiments.
Each experiment took about 1.5hrs, with half an hour per epoch. 
We also attempted to perform a grid search for BERT experiments, but it was too expensive to continue; therefore, we abandoned those jobs. 
Since we only used the \texttt{base} models, they amount to 110 million parameters. During fine-tuning, we do not freeze any parts of the model.

For CNN and PCNN, we performed grid search with Adam \cite{DBLP:journals/corr/KingmaB14} optimizer using learning rate $0.001$ for $20$ epochs with: batch size $\in \{128, 256\}$, bag size $\in \{4, 8, 16, 32\}$, 200-d word embeddings $\in$ $\{$Word2Vec \cite{mikolov2013distributed}\footnote{We used domain-specific word embeddings \emph{biowordvec} and \emph{biow2v} following \citet{marchesin-silvello-2022}.}, GloVe \cite{pennington2014glove}$\}$, and with (test-time) pooling $\in$ $\{ \text{ONE}, \text{AVG} \}$ when using sentence-level training and pooling in $\{ \text{ONE}, \text{AVG}, \text{ATT} \}$ when using bag-level training.
We ran this job on a cluster with support for array jobs. These amounted to over 700 experiments and took 3 days. 
We fixed other hyperparameters from literature \cite{han2018neural}, with position dimension set to 5, kernel size set to 3, and dropout set to 0.5. 
These are also default in OpenNRE \cite{han-etal-2019-opennre}.
The hyperparameters that had the most influence were batch size, bag size, and pre-trained word embeddings. 
All the experiments reported in this work are with a single run.

 \begin{table*}
    \centering
    \resizebox{15cm}{!}{
    \begin{tabular}{llll}
        \toprule
        {\bf Semantic Type} & {\bf 10k-20k} & {\bf 20k-30k} & {\bf $\geq$ 30k} \\
        \midrule
        Body Part, Organ, or Organ Component & \emph{bladder, heart, retinal, lungs, spinal, kidneys, colon} & \emph{eyes, lung, kidney, intestinal} & \emph{liver, brain} \\
        Organism Function & \emph{death} & \emph{period, blood pressure} & - \\
        Body Location or Region & \emph{head} & - & - \\
        \multirow{3}{*}{Therapeutic or Preventive Procedure} & \emph{injection, prevention, chemotherapy, application} & \emph{stimulation, delivery} & \emph{intervention, procedure, removal, operation} \\
        & \emph{resection, infusion, treatments, therapeutic} & & \\
        & \emph{surgical treatment, CT, surgical, transplantation} & & \\
        Neoplastic Process & \emph{cancer} & - & \emph{tumor, tumors} \\
        Disease or Syndrome & \emph{obesity, disorder, disorders} & \emph{diseases, stroke} & \emph{disease, infection, condition, hypertension} \\
        Laboratory Procedure & \emph{test, erythrocytes} & - & \emph{cells} \\
        Diagnostic Procedure & \emph{US, biopsy, ultrasound} & \emph{MRI} & - \\
        Finding & \emph{lesion, interaction, mass, difficulty, dependent} & \emph{abnormal} & \emph{presence, positive, negative, severe, lesions} \\
        Hormone & \emph{insulin} & - & - \\
        Biologically Active Substance & \emph{amino acids, glucose, ATP} & \emph{protein, proteins} \\
        Pharmacologic Substance & \emph{medication} & - & \emph{drugs, drug} \\
        Injury or Poisoning & \emph{strains} & \emph{injury, exposure} & \emph{damage} \\
        Tissue & \emph{tissue, bone marrow, tissues} & - & - \\
        Organism Attribute & \emph{male} & - & \emph{temperature, age} \\
        Immunologic Factor & \emph{antibody, antibodies} & - & - \\
        Health Care Activity & \emph{investigations} & \emph{examination} & \emph{assessment} \\
        Body Substance & \emph{plasma, blood, skin} & - & - \\
        Body System & - & \emph{cardiovascular} & - \\
        Mental Process & - & - & \emph{concentrations, concentration} \\
        Congenital Abnormality & - & \emph{abnormalities} & - \\
        \bottomrule
    \end{tabular}
    }
    \caption{Semantic types affected by type-based mention pruning with removed mentions placed in their respective frequency bins as discussed in \cref{section:knowledge_to_text}.}
    \label{table:pruned_mentions}
 \end{table*}

For sentence tokenization with ScispaCy, it took 9hrs with 32 CPUs (4GB each) and a batch size of 1024 to extract 151M sentences.
Further, the ScispaCy entity linking job took about half TB of RAM with 72 CPUs (6GB each) with a batch size of 4096 with 40hrs of run-time to link 145M unique sentences.

\begin{table*}
    \centering
    \begin{tabular}{C{7cm}C{7cm}}
    \toprule
    {\bf Relation} & {\bf Inverse Relation} \\
    \midrule
        \textit{finding\_site\_of} & \textit{has\_finding\_site} \\
        \textit{associated\_morphology\_of} & \textit{has\_associated\_morphology} \\
        \textit{method\_of} & \textit{has\_method} \\
        \textit{interprets} & \textit{is\_interpreted\_by} \\
        \textit{direct\_procedure\_site\_of} & \textit{has\_direct\_procedure\_site} \\
        \textit{causative\_agent\_of} & \textit{has\_causative\_agent} \\
        \textit{active\_ingredient\_of} & \textit{has\_active\_ingredient} \\
        \textit{interpretation\_of} & \textit{has\_interpretation} \\
        \textit{component\_of} & \textit{has\_component} \\
        \textit{indirect\_procedure\_site\_of} & \textit{has\_indirect\_procedure\_site} \\
        \textit{direct\_morphology\_of} & \textit{has\_direct\_morphology} \\
        \textit{cause\_of} & \textit{due\_to} \\
        \textit{direct\_substance\_of} & \textit{has\_direct\_substance} \\
        \textit{uses\_device} & \textit{device\_used\_by} \\
        \textit{focus\_of} & \textit{has\_focus} \\
        \textit{direct\_device\_of} & \textit{has\_direct\_device} \\
        \textit{procedure\_site\_of} & \textit{has\_procedure\_site} \\
        \textit{uses\_substance} & \textit{substance\_used\_by} \\
        \textit{associated\_finding\_of} & \textit{has\_associated\_finding} \\
        \textit{occurs\_after} & \textit{occurs\_before} \\
        \textit{is\_modification\_of} & \textit{has\_modification} \\
    \bottomrule
    \end{tabular}
    \caption{\textit{(Left)} 21 relations included in \meddistant, excluding NA relation. \textit{(Right)} For completeness, we also include their inverse relations.}
    \label{tab:rel_set_md19}
\end{table*}

\begin{table*}
    \centering
    \resizebox{7.0cm}{!}{
    \begin{tabular}{ccc}
        \toprule
        {\bf SG} & {\bf TUI} & {\bf Semantic Type} \\
        \midrule
        \multirow{8}{*}{ANAT} & T017 & Anatomical Structure \\
        & T029 & Body Location or Region \\
        & T023 & Body Part, Organ, or Organ Component \\
        & T030 & Body Space or Junction \\
        & T031 & Body Substance \\
        & T022 & Body System \\
        & T021 & Fully Formed Anatomical Structure \\
        & T024 & Tissue \\
        \midrule
        \multirow{17}{*}{CHEM} & T116 & Amino Acid, Peptide, or Protein \\
        & T195 & Antibiotic \\
        & T123 & Biologically Active Substance \\
        & T103 & Chemical \\
        & T200 & Clinical Drug \\
        & T196 & Element, Ion, or Isotope \\
        & T126 & Enzyme \\
        & T131 & Hazardous or Poisonous Substance \\
        & T125 & Hormone \\
        & T129 & Immunologic Factor \\
        & T130 & Indicator, Reagent, or Diagnostic Aid \\
        & T197 & Inorganic Chemical \\
        & T114 & Nucleic Acid, Nucleoside, or Nucleotide \\
        & T109 & Organic Chemical \\
        & T121 & Pharmacologic Substance \\
        & T192 & Receptor \\
        & T127 & Vitamin \\
        \midrule
        \multirow{2}{*}{DEVI} & T074 & Medical Device \\
        & T075 & Research Device \\
        \midrule
        \multirow{11}{*}{DISO} & T020 & Acquired Abnormality \\
        & T190 & Anatomical Abnormality \\
        & T049 & Cell or Molecular Dysfunction \\
        & T019 & Congenital Abnormality \\
        & T047 & Disease or Syndrome \\
        & T033 & Finding \\
        & T037 & Injury or Poisoning \\
        & T048 & Mental or Behavioral Dysfunction \\
        & T191 & Neoplastic Process \\
        & T046 & Pathologic Function \\
        & T184 & Sign or Symptom \\
        \midrule
        \multirow{6}{*}{PHYS} & T201 & Clinical Attribute \\
        & T041 & Mental Process \\
        & T032 & Organism Attribute \\
        & T040 & Organism Function \\
        & T042 & Organ or Tissue Function \\
        & T039 & Physiologic Function \\
        \midrule
        \multirow{7}{*}{PROC} & T060 & Diagnostic Procedure \\
        & T065 & Educational Activity \\
        & T058 & Health Care Activity \\
        & T059 & Laboratory Procedure \\
        & T063 & Molecular Biology Research Technique \\
        & T062 & Research Activity \\
        & T061 & Therapeutic or Preventive Procedure \\
        \bottomrule
    \end{tabular}
    }
    \caption{51 semantic types (STY) along with their TUIs and semantic groups (SG) covered in \textsc{MedDistant19}.}
    \label{table:md19_semantic_types_and_groups}
 \end{table*}

\end{document}